\documentclass[lettersize,journal,twoside]{IEEEtran}
\usepackage{amsmath,amsfonts}
\usepackage{algorithmic}
\usepackage{array}
\usepackage[caption=false,font=normalsize,labelfont=sf,textfont=sf]{subfig}
\usepackage{textcomp}
\usepackage{stfloats}
\usepackage{url}
\usepackage{verbatim}
\usepackage{graphicx}
\def\BibTeX{{\rm B\kern-.05em{\sc i\kern-.025em b}\kern-.08em
    T\kern-.1667em\lower.7ex\hbox{E}\kern-.125emX}}
\usepackage{balance}

\usepackage{caption}
\usepackage{adjustbox} 
\usepackage{multirow} 
\usepackage{booktabs}
\usepackage{threeparttable}
\usepackage{makecell}
\usepackage{float}

\usepackage{fancyhdr}

\IEEEoverridecommandlockouts                     

\begin{document}
\title{\LARGE \bf
Development of the Bioinspired Tendon-Driven DexHand 021 with Proprioceptive Compliance Control}

\author{Jianbo YUAN $^{1}$  Haohua ZHU $^{2}$ Jing DAI $^{1}$ and Shen YI $^{2}$

\thanks{Manuscript received: July 16, 2025; Revised: September 30, 2025; Accepted: October 26, 2025. This paper was recommended for publication by Editor Xinyu Liu upon evaluation of the Associate Editor and Reviewers’ comments. This research supported by  DEXROBOT CO., LTD. (Corresponding author: Jianbo YUAN)}%
\thanks{$^{1}$The first and third authors are with the School of Mechanical Engineering, Shanghai Jiao Tong University, Shanghai 200240, China.
        {\tt\small jianbo.yuan@sjtu.edu.cn; daisy\_dai@sjtu.edu.cn}}%
\thanks{$^{2}$The second and fourth authors are with DEXROBOT CO., LTD., Zhejiang 312455, China.
        {\tt\small zhh@dex-robot.com; ys@dex-robot.com}}%
\thanks{Digital Object Identifier (DOI): see top of this page.}
}

\maketitle

\begin{abstract}
The human hand plays a vital role in daily life and industrial applications, yet replicating its multifunctional capabilities-including motion, sensing, and coordinated manipulation with robotic systems remains a formidable challenge. Developing a dexterous robotic hand requires balancing human-like agility with engineering constraints such as complexity, size-to-weight ratio, durability, and forcesensing performance. This letter presents DexHand 021, a high-performance, cable-driven five-finger robotic hand with 12 active and 7 passive degrees of freedom (DOFs), achieving 19 DOFs dexterity in a lightweight 1 \(kg\) design. We propose a proprioceptive force-sensing-based admittance control method to enhance manipulation. Experimental results demonstrate its superior performance: a single-finger load capacity exceeding 10 \(N\), fingertip repeatability under 0.001 \(m\), and force estimation errors below 0.2 \(N\). Compared to PID control, joint torques in multi-object grasping are reduced by 31.19 \(\%\), significantly improves force-sensing capability
while preventing overload during collisions. The hand excels in both power and precision grasps, successfully executing 33 GRASP taxonomy motions and complex manipulation tasks. This work advances the design of lightweight, industrialgrade dexterous hands and enhances proprioceptive control, contributing to robotic manipulation and intelligent manufacturing.
\end{abstract}


\section{INTRODUCTION}

Dexterous manipulation presents fundamental challenges in robotics. While human hands demonstrate exceptional grasping and tool manipulation abilities essential for industrial and daily tasks, replicating this dexterity in robotic systems remains an active research frontier. The development of anthropomorphic robotic hands represents a key interdisciplinary challenge spanning robotics, intelligent manufacturing, and machine learning ~\cite{1,2}.

The human hand excels in motion and manipulation due to muscle-driven actuation, but artificial systems struggle with flexibility and size-to-weight ratios. Motor-cable transmission offers human-like dexterity, yet placing actuators in the forearm increases size, weight, and complexity. Examples like the Shadow-hand ~\cite{3} and NASA Robonant-2-hand ~\cite{4} emulate up to 22 DOFs (20 independently controlled DOFs) achieving high dexterity. Nevertheless, complex systems are often hindered by three key challenges: excessive weight, reduced operational lifespan, and control inefficiencies. The Shadow-hand demonstrates human-like dexterity; however, its 4.3 \(kg\) mass, short cable service life, and intricate control logic hinder broad deployment. The TRX-hand-5 ~\cite{5} reduces actuators but remains heavy at 2.6 \(kg\), failing to resolve cable transmission issues. Rigid mechanisms (e.g., linkages and gears) provide high control precision, but their inherent design constraints limit dexterity. Representative examples like the DLR Hand ~\cite{6} and Schunk Hand ~\cite{7} exhibit structural complexity that renders them impractical for mass production. Pneumatic actuation, like Bionic-soft-hand ~\cite{8} and SJTU-MIT-hand ~\cite{9}, mimics muscles but lacks precision and compactness, failing to meet industrial stability and precision needs. Other underactuated hands ~\cite{1,2,3} struggle to balance DOFs, weight, compliance, and sensing capabilities.

\begin{figure}[t]
	\centering
	\includegraphics[width=0.8\linewidth]{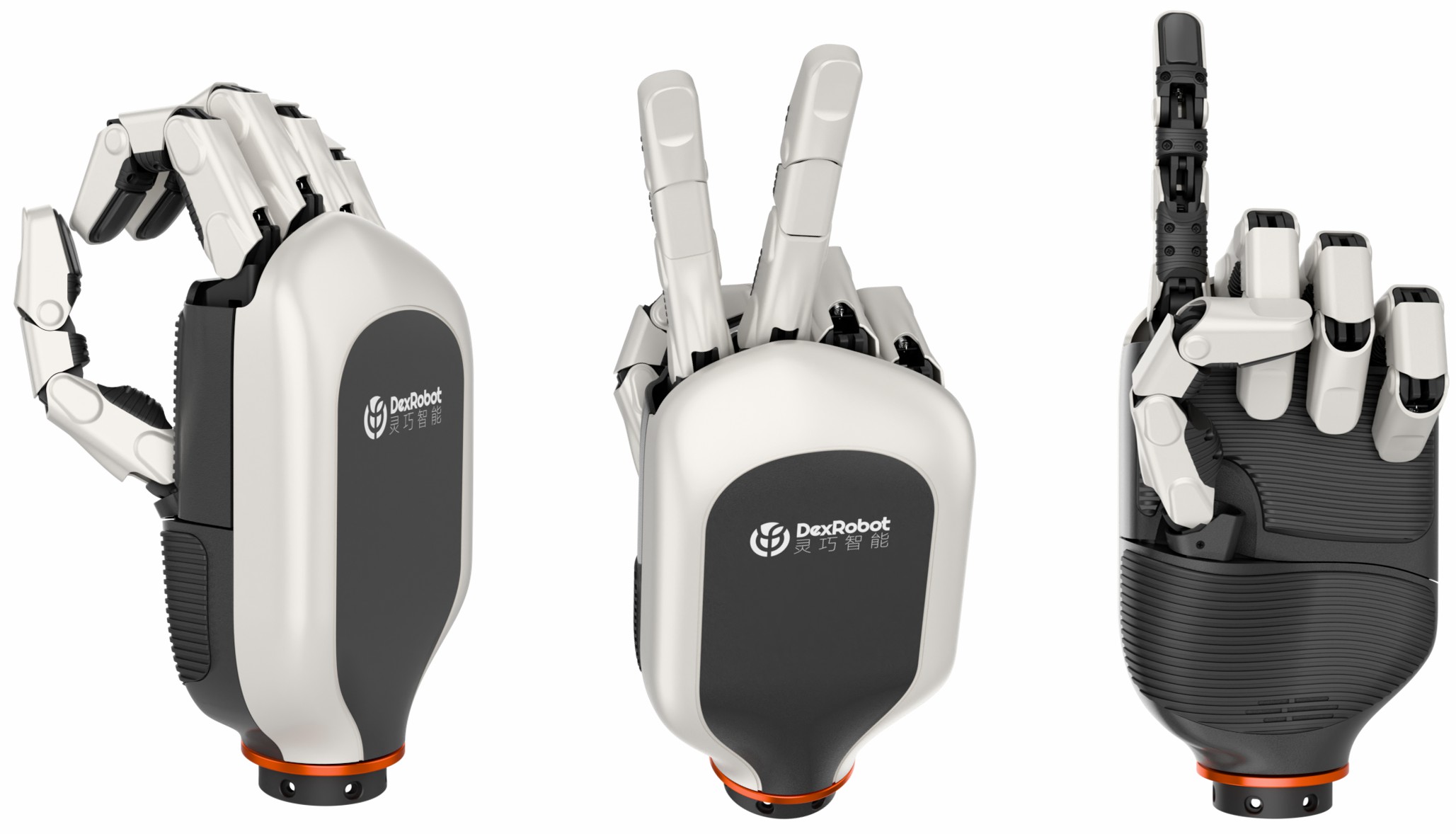}
        \captionsetup{font=footnotesize, labelfont=footnotesize}
	\caption{DexHand 021}
\end{figure}

Recent advancements in robotic hands have focused on integrating tactile sensing, a field gaining significant attention ~\cite{10}. While force conversion mechanisms are well understood, embedding sensors, circuits, and communication devices into compact systems remains challenging ~\cite{11}. Common force-sensing methods rely on resistance, capacitance, or visual signals, but miniaturizing these components, routing wiring across joints, and developing efficient signal processing algorithms are ongoing hurdles ~\cite{12,13}. Full-hand tactile sensing, akin to human skin, depends on sensing technology miniaturization. Notable examples include the Robonaut-2-hand ~\cite{4} with multidirectional load sensors, the Gifu hand-III ~\cite{14} with piezoresistive arrays, the Shadow-hand ~\cite{3} using Bio-Tac sensors, the Allegro-hand ~\cite{10} employing U-skin, and Gelslim  ~\cite{15} with vision-based tactile sensing. Enhancing perception via sensors or algorithms is critical for better manipulation. When full-hand sensing is impractical, joint torque estimation via algorithms offers a viable alternative. Motor-cable transmission shows promise but introduces nonlinear challenges like elastic deformation, friction, and hysteresis, impacting control precision ~\cite{16,17}. Existing research on cable-driven systems addresses cable-specific issues but offers limited improvements for high-DOFs devices like dexterous hands.

Inspired by human hand biomechanics, a coupled physical-informational model was developed to guide high-performance anthropomorphic end-effector design. Computational muscle models integrated with artificial actuators simulate physiological contraction dynamics, yielding bio-inspired strategies. The DexHand 021 prototype demonstrates this approach, achieving dexterity and compliance via muscle-mimetic structures. To bypass force measurement constraints, statistical joint torque estimation models embed simplified muscular dynamics into actuator control, emulating biological behaviors. Experiments validate enhanced robustness and adaptive control through admittance-based grasping.

The paper is structured as follows: Chapter 2 details DexHand 021’s design; Chapter 3 introduces torque estimation and control; Chapter 4 validates experiments; the final chapter concludes and outlines future work.

\begin{figure}[t]
	\centering
	\includegraphics[width=0.9 \linewidth]{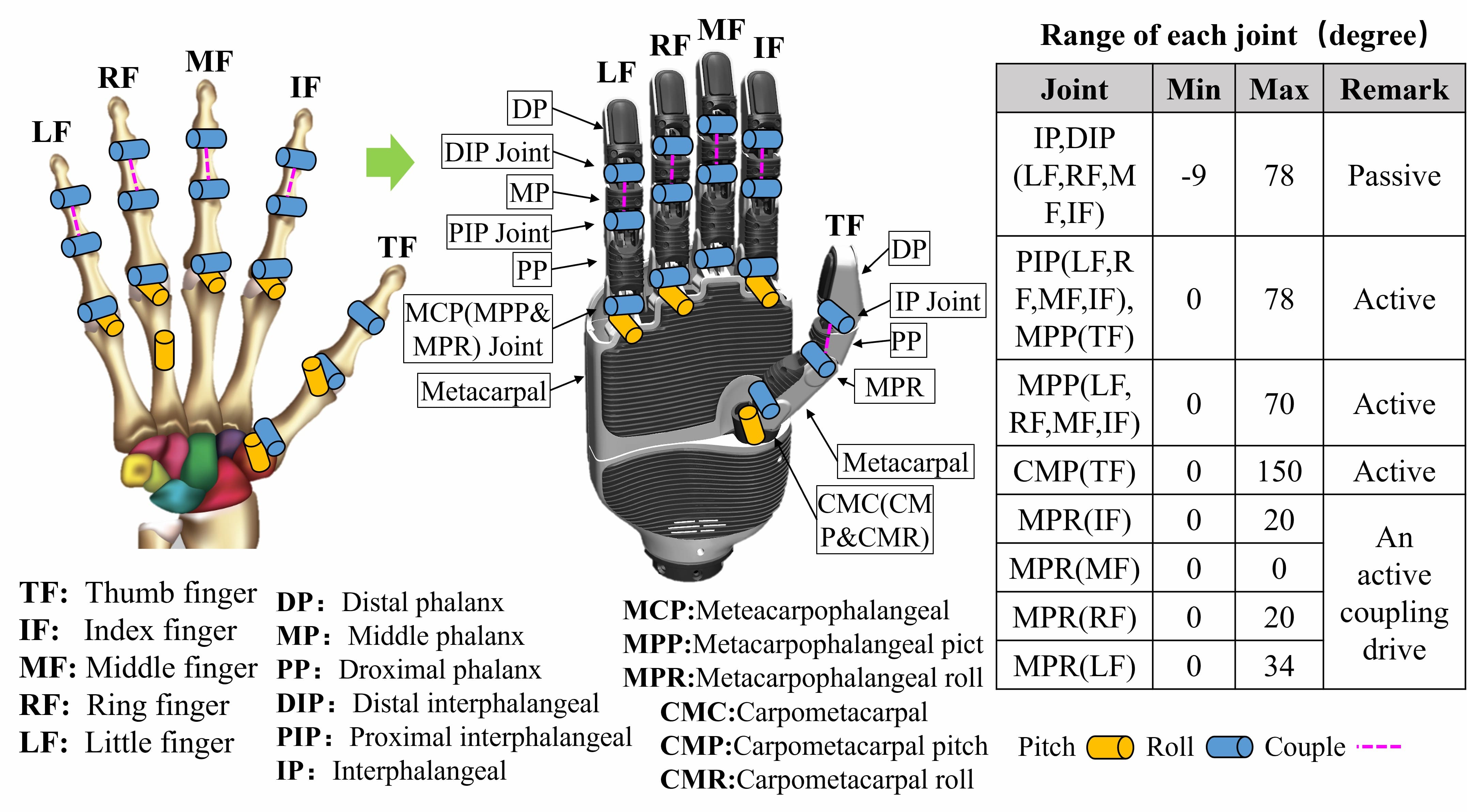}
        \captionsetup{font=footnotesize, labelfont=footnotesize}
	\caption{DOF design of DexHand 021}
\end{figure}

\section{DESIGN OF THE RECONFIGURABLE DexHand 021}

\subsection{DexHand 021 DOF Design}

The human hand comprises 19 bones driven by 29 muscles, offering 22 DOFs crucial for manipulation. Studies indicate that the thumb, index, and middle fingers can perform 29 of 33 common grasping tasks, while tasks like using scissors or chopsticks are more complex ~\cite{18}. Replicating the hand’s full morphology is challenging due to motor constraints. To balance dexterity, weight, maintainability, and cost, a design with 12 brushless hollow-cup motors driving 19 DOFs was developed. This includes thumb, index, middle, ring, and little fingers, each with DIP (underactuated), PIP, MPP, and MPR joints, plus additional CMP and CMR for the thumb. The hardware setup and DOFs design are detailed in Figs.1 and 2.

\subsection{Artificial Muscle Design}

Muscle structure comprises fibers and tendons, with fibers aligned at a pennation angle to the tendon (Fig.3(a)). Activation signals influence tendon force by altering fiber length and contraction velocity. External forces change fiber length, generating passive forces through elastic components. In 1938, Hill ~\cite{19} proposed a Hill-type numerical model with three elements: contractile, parallel elastic, and series elastic. Tendon force, length, and contraction velocity exhibit strong nonlinearity (Fig.3(b, c)). This model is foundational for simulating muscle dynamics in biomechanics and robotics. In this study, muscle contraction is simulated using a DC motor-cable drive mechanism, with the \(\alpha = 0\)  (Fig.3(d, e, f)). The control signal is represented by current, and a linear approximation is used to derive Equation (1). Muscle fiber force is calculated using the following equation:

\begin{align}
F^m = & I + e^{K^p(\hat{l} - l)} + \notag \\
      & \left(K^{d1}I + K^{d2}\right)(\hat{\dot{l}} - \dot{l}) + K^s(l^s - l^m)
\end{align}
where, \(F^m\) represents tendon force; \(l\) and  \(\hat{l}\) represent the actual cable length and the desired cable length, respectively; \(\dot{l}\) and \(\hat{\dot{l}}\) represent the actual cable velocity and the desired cable velocity, respectively; \(l^s\)  and \(l^m\) represent the actual length of the passive spring and the initial preloaded length, respectively; \(K^p,K^{d1},K^{d2}\) and \(K^s\) represent positive definite coefficient.

 \begin{figure}[t]
	\centering
	\includegraphics[width=0.9\linewidth]{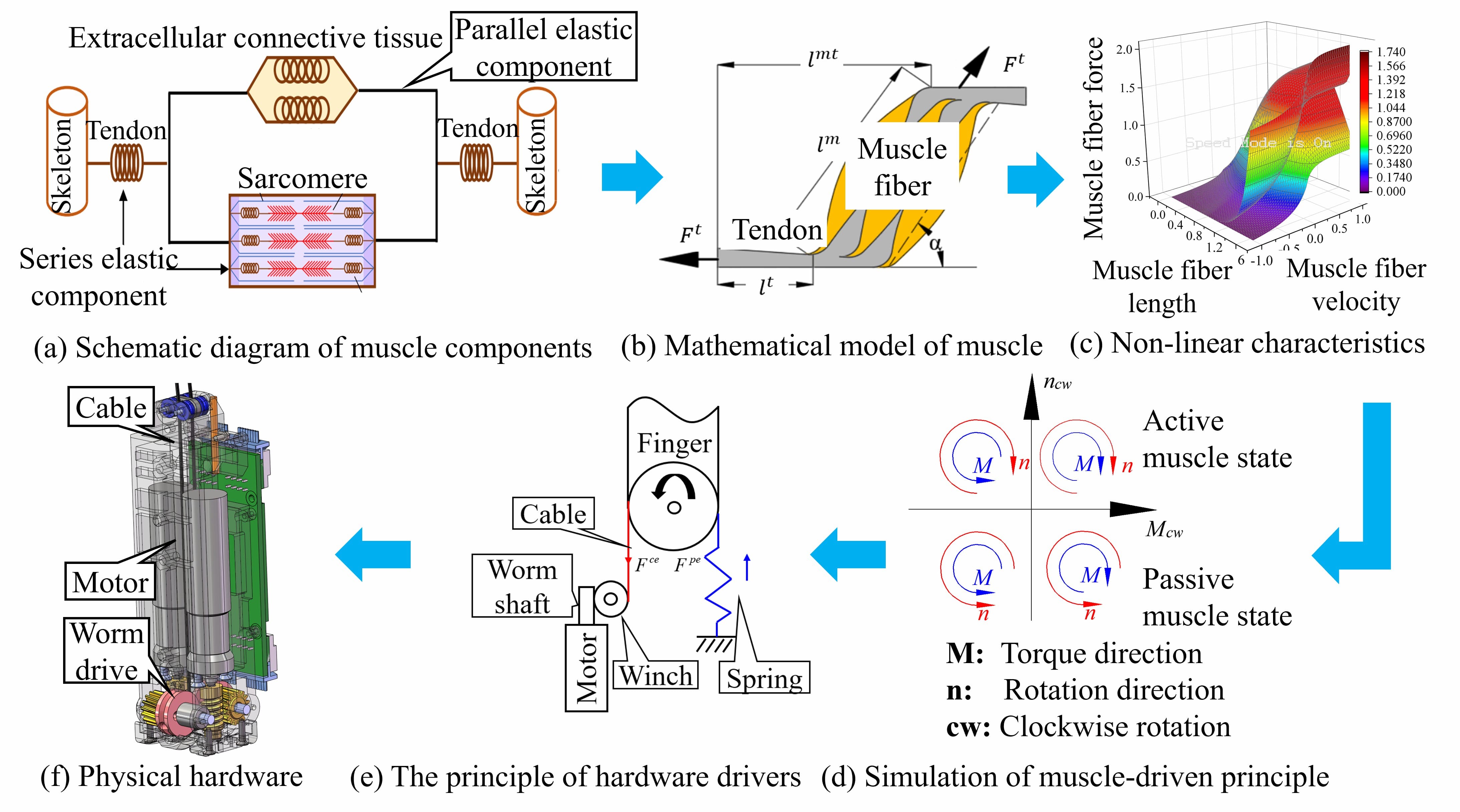}
        \captionsetup{font=footnotesize, labelfont=footnotesize}
	\caption{Biological mechanisms to muscle-like physical realization: Utilizing the driving characteristics of motors and cable to simulate muscle function, Quadrant I simulates active contraction; Quadrant IV simulates passive contraction.}
\end{figure}

 \begin{figure}[t]
	\centering
	\includegraphics[width=0.9\linewidth]{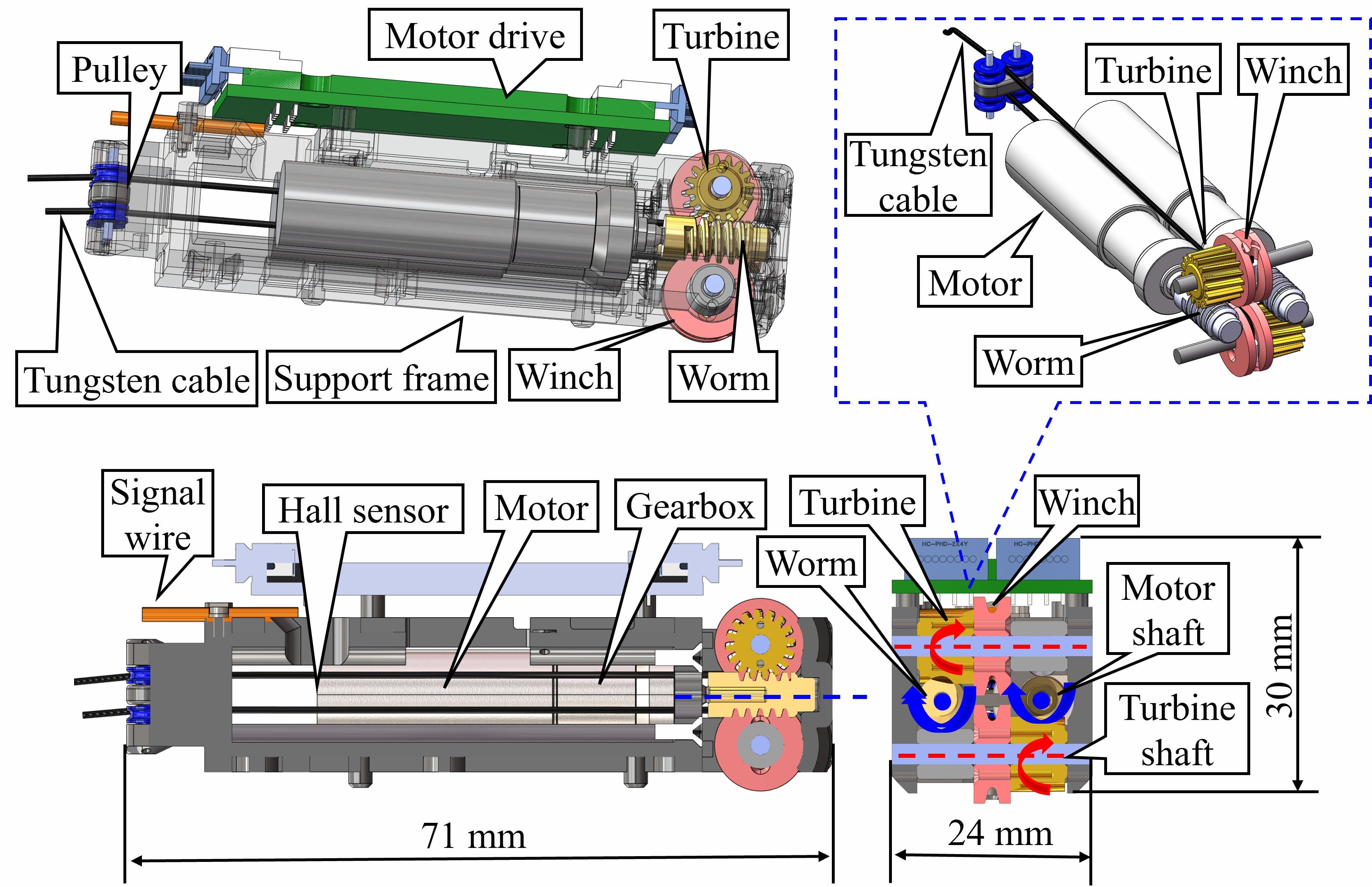}
        \captionsetup{font=footnotesize, labelfont=footnotesize}
	\caption{Dual-drive artificial muscle design: Hollow cup motors drive worm gears, pulling cords for joint contraction/relaxation.}
\end{figure}

\begin{figure}[t]
	\centering
	\includegraphics[width=0.9\linewidth]{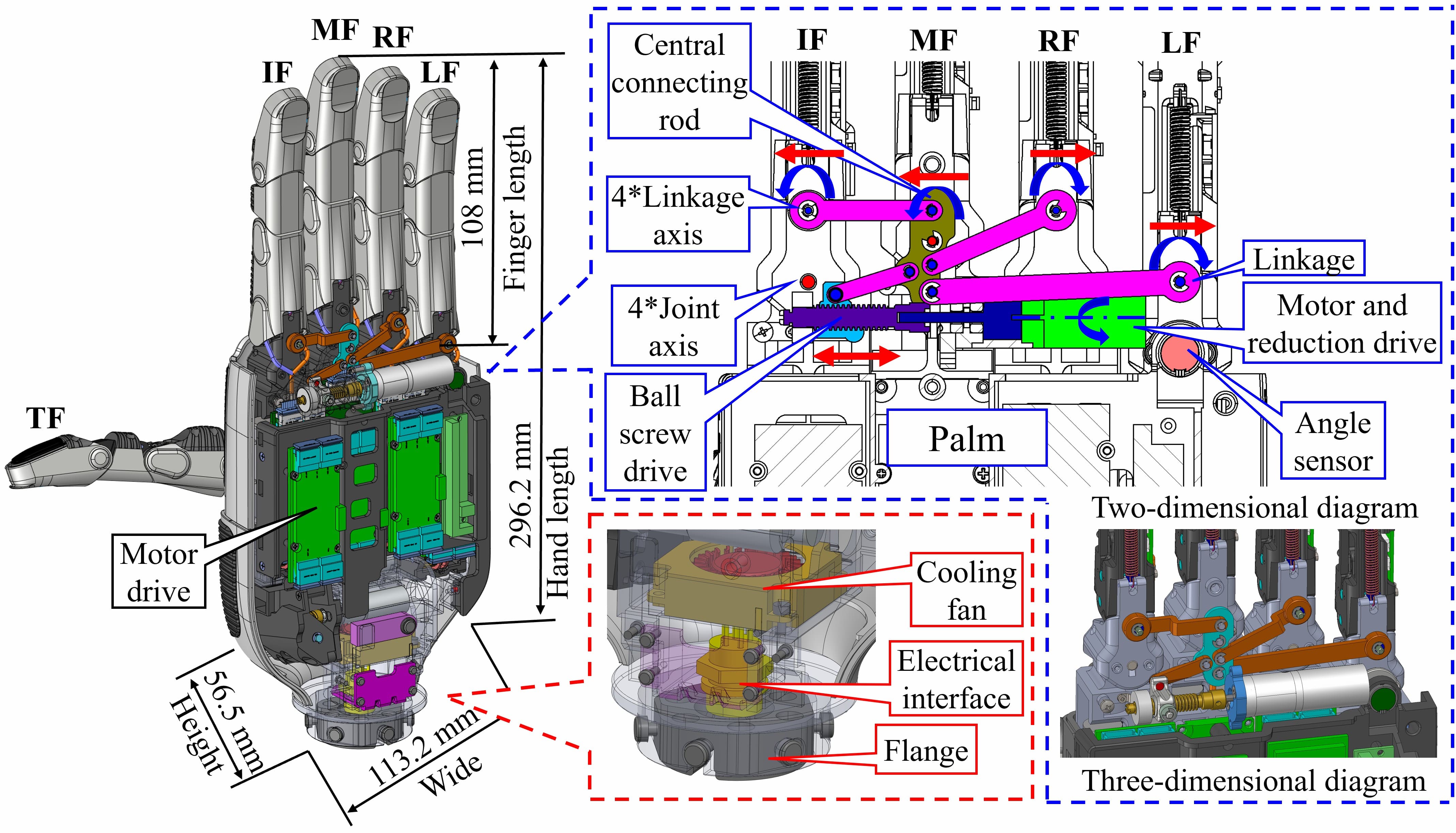}
        \captionsetup{font=footnotesize, labelfont=footnotesize}
	\caption{Size, finger division and heat dissipation design: 1) Motor-driven lead screws convert rotation to linear motion, enabling independent finger movement via a crank-link mechanism. 2) An active cooling fan near the flange monitors driver and motor temperatures for efficient heat dissipation.}
\end{figure}

Simulating muscle-driven characteristics within confined spaces poses significant challenges. High-power-density hollow-cup DC motors, coupled with planetary reducers and worm gear mechanisms, were selected to overcome the limitations of miniature reducers' terminal gears and enhance static locking capability. Two motors were integrated into a single bracket using an efficient design to optimize space utilization. The bracket, fabricated from high-precision machined plastic, offers simplicity in manufacturing, low cost, and high strength. Other core components are made from high-precision machined copper or aluminum alloys. The design schematic is shown in Fig.4. The transmission pulleys are designed to retain lubrication despite their compact size. Multi-braided, soft, high-strength tungsten cables (diameter: 0.76 \(mm\), tensile strength \(>\) 650 \(N\)) were chosen for the tendons, providing minimal strain, high wear resistance, and excellent strength, ensuring stable and reliable force transmission. Each artificial muscle unit consumes approximately 6 \(W\) of power and, with the reduction and capstan mechanisms, can continuously output a pulling force of 150 N. According to Equation (1), the contraction characteristics of muscles are integrated into the artificial muscle unit, The physical implementation scheme is shown in Fig.4.

\begin{figure}[t]
	\centering
	\includegraphics[width=0.9\linewidth]{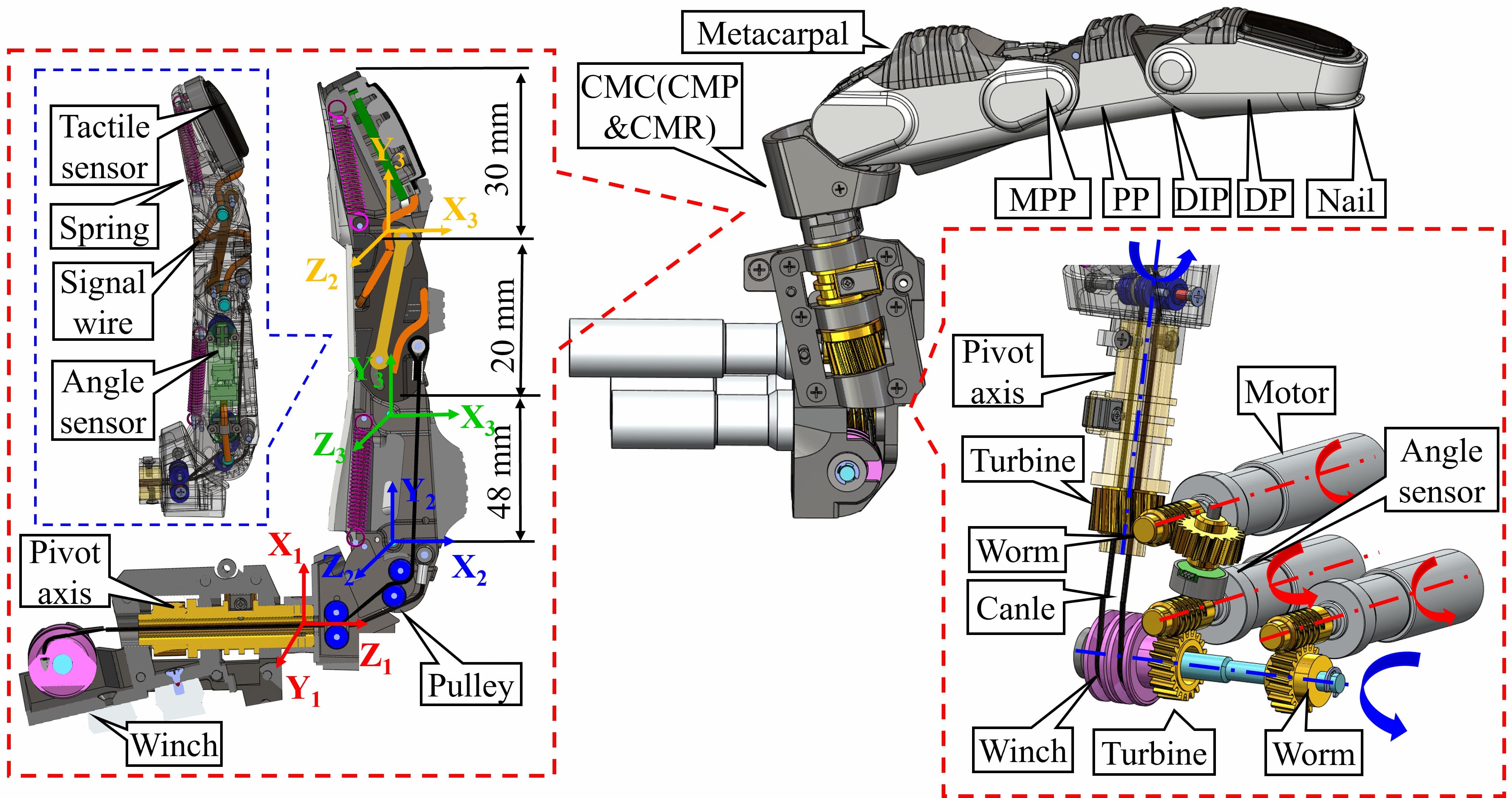}
        \captionsetup{font=footnotesize, labelfont=footnotesize}
	\caption{Modular design of the thumb: The three-motor-tendon system actuates four joints through the CMP joint axis.}
\end{figure}

\begin{figure}[t]
	\centering
	\includegraphics[width=0.9\linewidth]{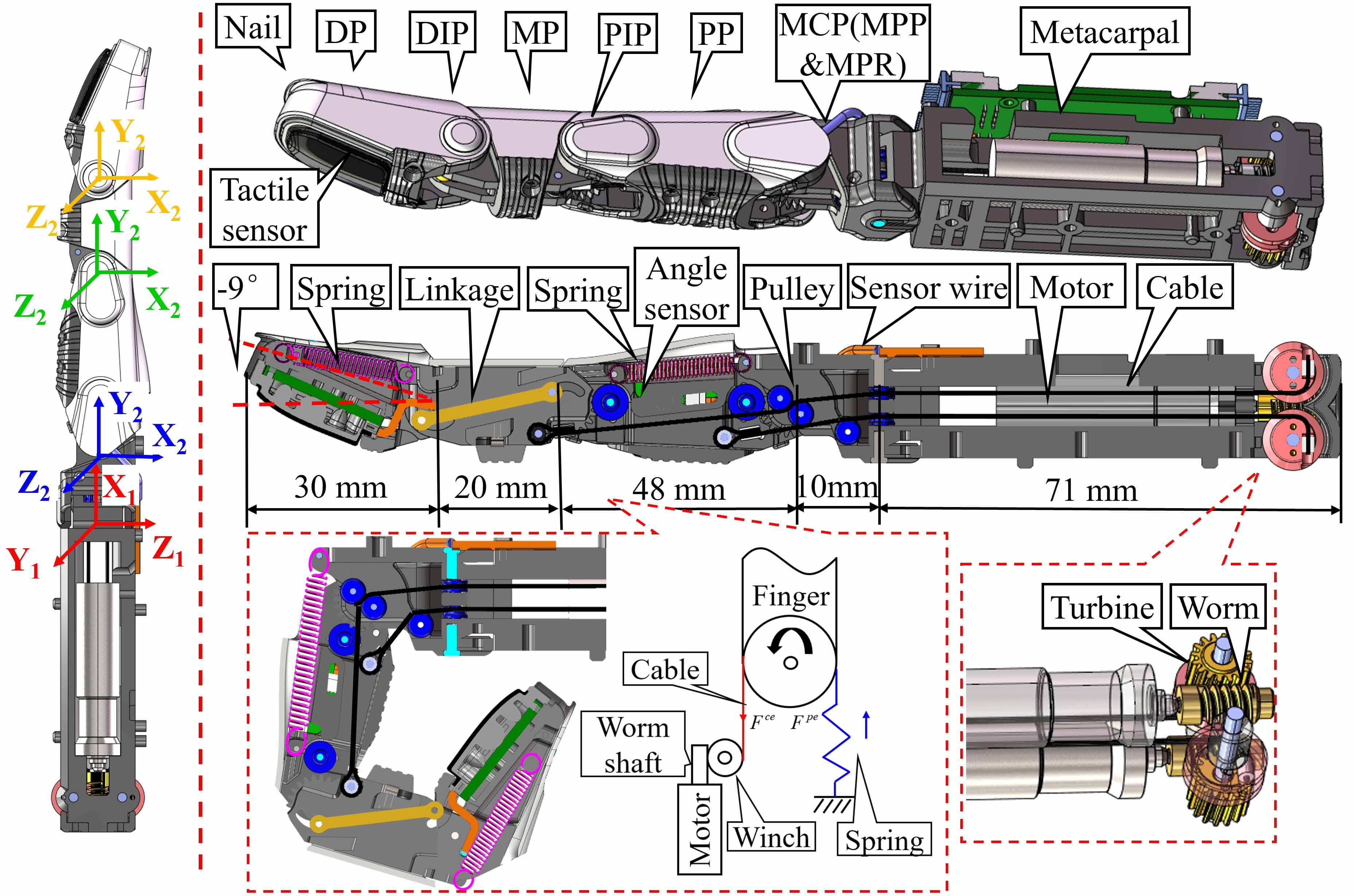}
        \captionsetup{font=footnotesize, labelfont=footnotesize} 
	\caption{Modular design of the four-finger: Two motor-tendon systems drive three joints.}
\end{figure}

\begin{figure}[t]
	\centering
	\includegraphics[width=0.9\linewidth]{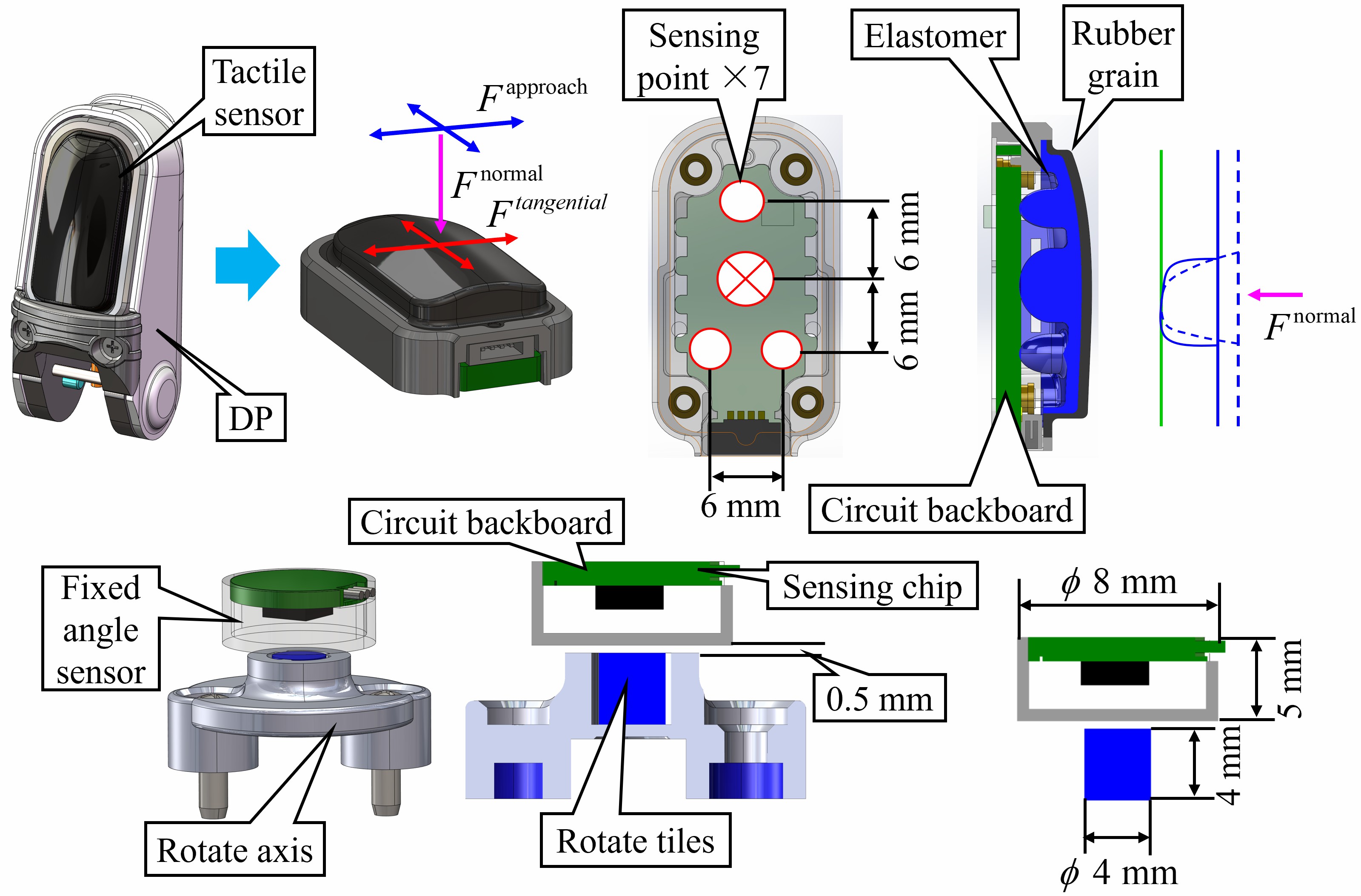}
        \captionsetup{font=footnotesize, labelfont=footnotesize}
	\caption{ Principles of array multi-axis force and Hall angle sensors.}
\end{figure}

\begin{figure}[t]
	\centering
	\includegraphics[width=0.9\linewidth]{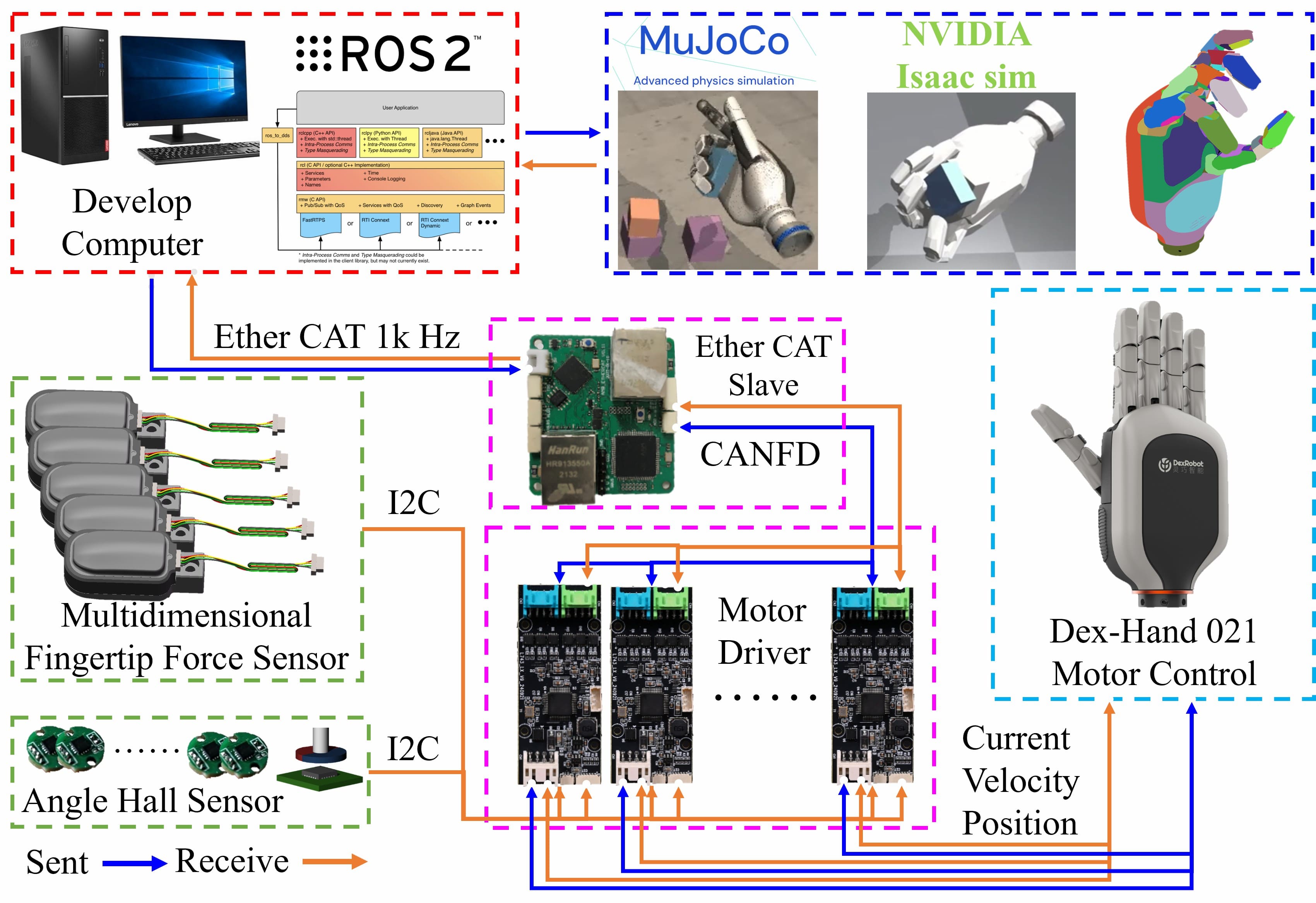}
        \captionsetup{font=footnotesize, labelfont=footnotesize}
	\caption{High-speed hard real-time embedded system and digital twin simulation platform: Connect the simulation environment with DexHand 021 hardware through the ROS2 system to achieve unified communication.}
\end{figure}

\subsection{Modular Finger Design}
The DexHand 021 mimics the human hand’s anatomy with an underactuated design to reduce size and weight while preserving key joint functionality. It consists of three modules: the underactuated MPR joints for the four fingers, the thumb module, and the four-finger module. While the thumb and fingers share a similar skeletal design, their joint layouts and actuation differ. A DC motor and screw-slider mechanism convert rotary motion into linear motion, driving the MCP joints via a multi-link structure that replicates human hand kinematics (Fig.5). The thumb, designed independently, uses a DC motor with a worm gear for CMR joint rotation. Two additional motors drive worm gears to actuate the CMP and MPP joints via tendons, avoiding contact with the CMR axis. The DIP and MPP joints employ an underactuated linkage, with the DIP joint angled at -9° for fingertip coordination. The thumb’s DP includes a nail structure for precision tasks (Fig.6). The four fingers share a modular design. A dual-row DC motor assembly with worm gears actuates tendons controlling the MPP and PIP joints. The DIP and PIP joints use underactuated linkages, with the DIP joint naturally resting at -9° for enhanced fingertip motion. The DP also features a nail structure for fine operations (Fig.7). Custom motor-tendon artificial muscle modules actuate all joints except the CMR, optimizing palm space. The design integrates actuation, transmission, sensing, and communication within 296.2 \(mm\) × 113.2 \(mm\) × 56.5 \(mm\), weighing 1 \(kg\).

\subsection{Sensing System Design}
The DexHand 021’s DP fingertips are equipped with multi-point capacitive tactile sensors, capable of detecting normal and tangential forces with accuracies of 0.1 \(N\) and 0.25 \(N\), respectively. Additionally, these sensors can detect dielectric constant signals of different materials within a 10 mm range. Seven sensing points, including four symmetrically positioned at the center, enable precise detection of force direction changes. Each actively controlled joint incorporates a Hall-effect angle sensor with a resolution of 0.1 °, ensuring accurate motion tracking in confined spaces (Fig.8). This design integrates high sensitivity, accuracy, and compactness, making it suitable for advanced robotic manipulation tasks.
\subsection{Miniaturized Integrated Embedded System Design}

This section tackles the challenge of high-density ‘mechanical-electrical-sensing-control’ integration in dexterous hands by proposing a compact driver based on an automotive-grade MCU. The driver optimizes hardware layout, mitigates signal interference, and supports multi-channel motor control and sensor data acquisition via I2C and ADC interfaces. It integrates dual BLDC motor control for torque regulation and overcurrent protection, while complying with the ISO26262 ASIL-B safety standard. With dimensions of 24 \(mm\) × 58 \(mm\) × 8 \(mm\), the driver enables hard real-time communication through CANFD (5 \(Mbps\)) and integrates with Ether-CAT, ROS, NVIDIA Isaac-Gym, and MUJOCO for enhanced system flexibility (Fig.9). The DexHand 021’s parameters are detailed in Tab.1.

\begin{table}[t]
        \centering
        \caption{DexHand 021 Parameters}
        \label{tab: DexHand 021 Parameters}
        \begin{tabular}{ll}
        \hline
        \textbf{Project} & \textbf{Parameters} \\
        \hline
        Dimension & 296.2 \(mm\) $\times$ 113.2 \(mm\) $\times$ 56.5 \(mm\) \\
        Weight & 1 \(kg\) \\
        Motor & 12 hollow-cup DC motors (3 \(W\)) \\
        Fingertip force & \begin{tabular}[c]{@{}l@{}}Manipulation finger: 50 \(N\) \\ Grasping finger: 10 \(N\) \end{tabular} \\
        Joint angle & \begin{tabular}[c]{@{}l@{}}Accuracy: $\pm 0.5^\circ$ \\ \end{tabular} \\
        Joint speed & 200$^\circ$/s \\
        Tactile sensors & \begin{tabular}[c]{@{}l@{}}Normal force: 0.1 \(N\)/20\(N\) \\ Tangential force: 0.25 \(N\) \\ Proximity sense: 10 \(mm\)\end{tabular} \\
        Communication & Ether CAT / CADFD / ROS / Python / C++ \\
        \hline
        \end{tabular}
\end{table}

\section{SELF-PERCEIVED COMPLIANT CONTROL}
\subsection{Nonlinear Compensation Algorithm} 

Gaussian Process Regression (GPR) is widely used for data modeling and uncertainty estimation ~\cite{20,21}. It employs kernel functions to model complex, nonlinear relationships in multi-feature data, particularly in high-dimensional spaces. GPR's uncertainty quantification improves reliability and supports data analysis, especially in small-sample, high-dimensional scenarios, reducing overfitting and ensuring robust predictions. The joints torque estimation of DexHand 021 is framed as a multivariate nonlinear fitting problem. GPR is applied here, using motor current, motor position, motor velocity, joint position, joint velocity, and temperature as inputs to predict measured joints torque. The learned mapping relationship \(f: X \to Y\)  is then used to predict new outputs corresponding \(X^*\) to new inputs \(Y^*\). The problem is first defined as:

\begin{equation}
    \begin{aligned}
        \mathbf{X} = [X_1, X_2, \ldots, X_N]^T, \quad X_i \in \mathbb{R}^d\\
        \mathbf{Y} = [Y_1, Y_2, \ldots, Y_N]^T,
         \quad Y_i \in \mathbb{R}
    \end{aligned}
\end{equation}

In GPR modeling, the kernel function is defined as:
\begin{equation}
    k(x, x') = k_1(x, x') + k_2(x, x') \tag{3}
\end{equation}
where \(k_1(x, x') = \sigma_f^2 \exp\left(-\|x - x'\|^2 / (2l^2)\right)\)  is the RBF kernel, representing the similarity between data points, and \(k_2(x, x') = \sigma_n^2 \delta(x, x')\)  is the white noise kernel, representing measurement noise.

Optimize the model by maximizing the marginal likelihood function to determine the hyperparameters \(\theta = \{\sigma_f, l, \sigma_n\}\):
\begin{align}
    \log p(Y | X, \theta) = & -\frac{1}{2} Y^T (K + \sigma_n^2 I)^{-1} Y \notag \\
                            & - \frac{1}{2} \log |K + \sigma_n^2 I| - \frac{n}{2} \log(2\pi) 
    \tag{4}
\end{align}

The model prediction \(Y_{\text{pred}}\) performed using the posterior mean   and standard deviation \(\sigma\) :

\begin{equation}
    Y_{\text{pred}} = \mu, \quad \sigma = \sqrt{\text{diag}(\Sigma)} \tag{5}
\end{equation}

\subsection{Impedance Control Based on Joint Torque Estimation}

\begin{figure}[t]
	\centering
	\includegraphics[width=0.9\linewidth]{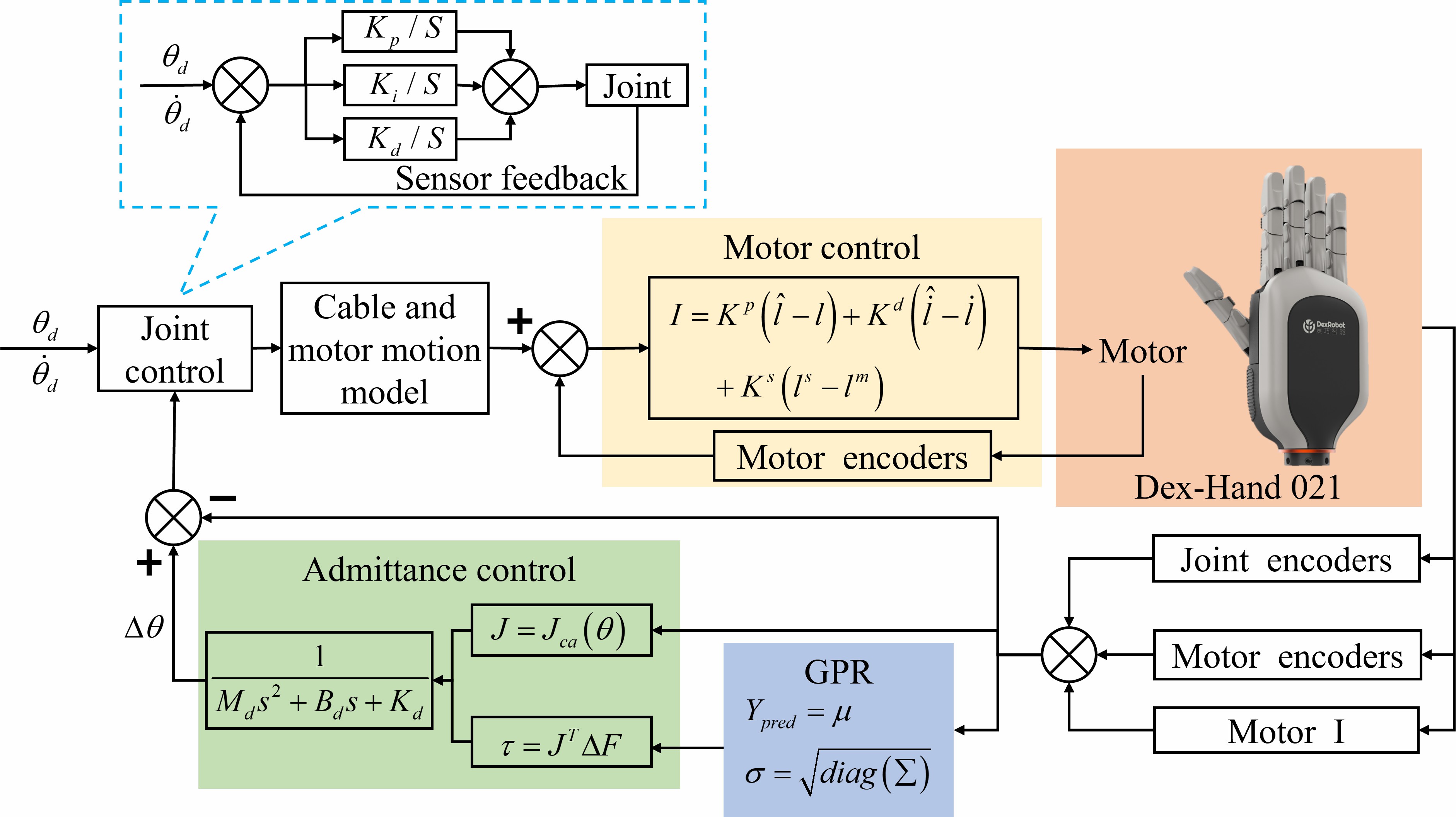}
        \captionsetup{font=footnotesize, labelfont=footnotesize} 
	\caption{Admittance control block diagram.}
\end{figure}

\begin{figure}[t]
	\centering
	\includegraphics[width=0.9\linewidth]{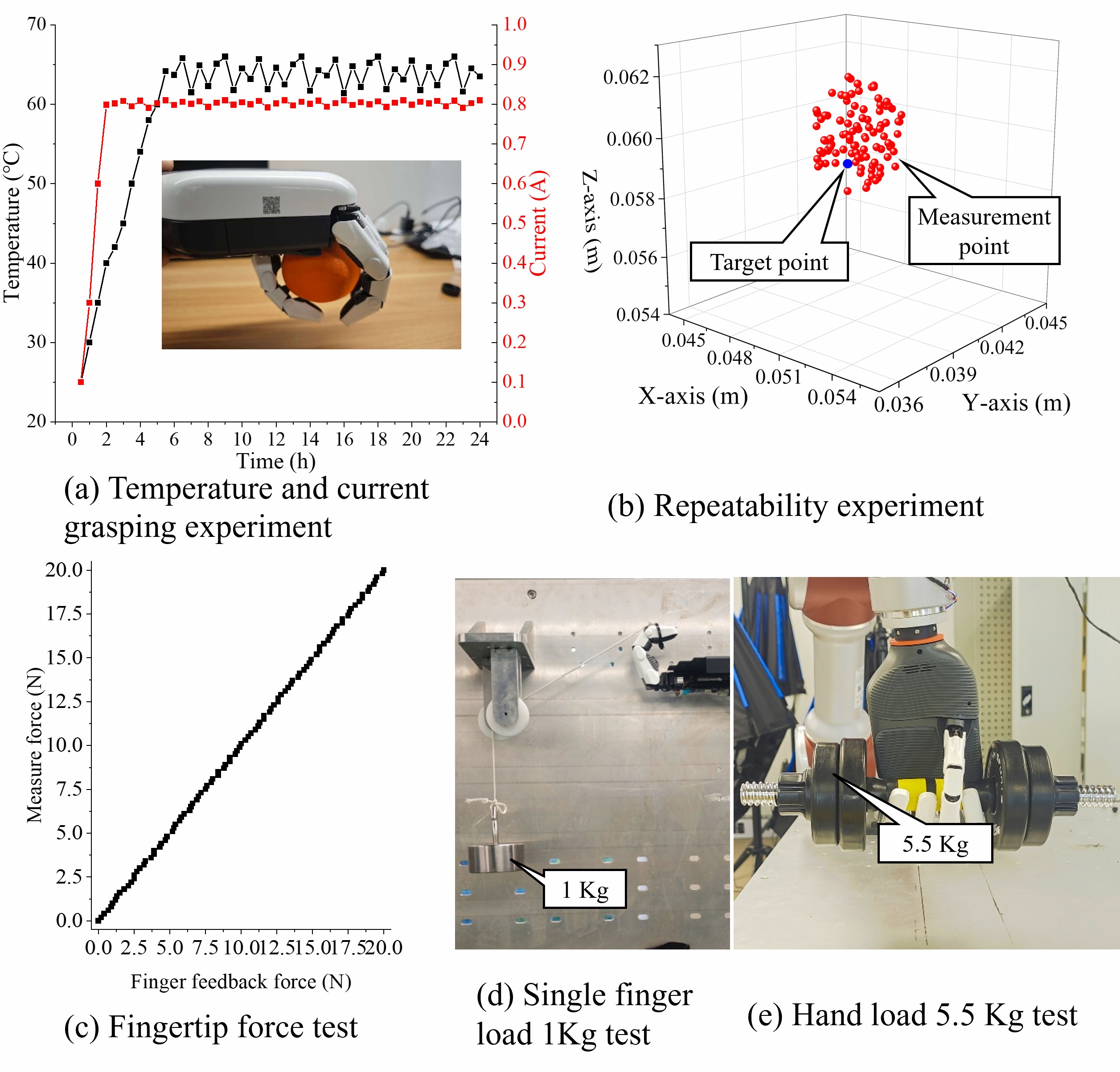}
        \captionsetup{font=footnotesize, labelfont=footnotesize} 
	\caption{DexHand 021 Performance Experiment.}
\end{figure}

\begin{figure}[t]
	\centering
	\includegraphics[width=0.9\linewidth]{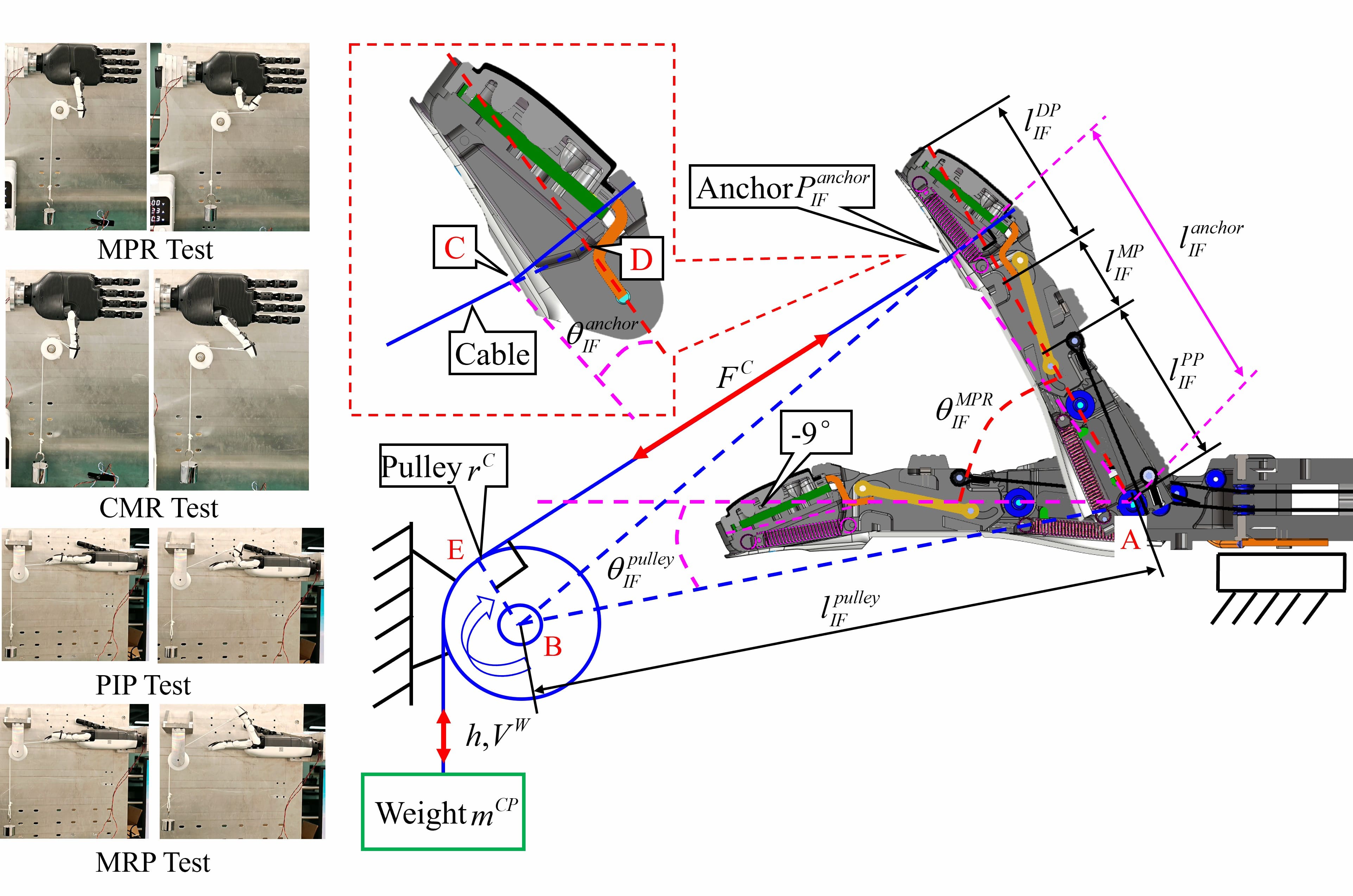}
        \captionsetup{font=footnotesize, labelfont=footnotesize} 
	\caption{Joint torque testing principles and experimental procedures.}
\end{figure}

\begin{figure}[t]
	\centering
	\includegraphics[width=0.9\linewidth]{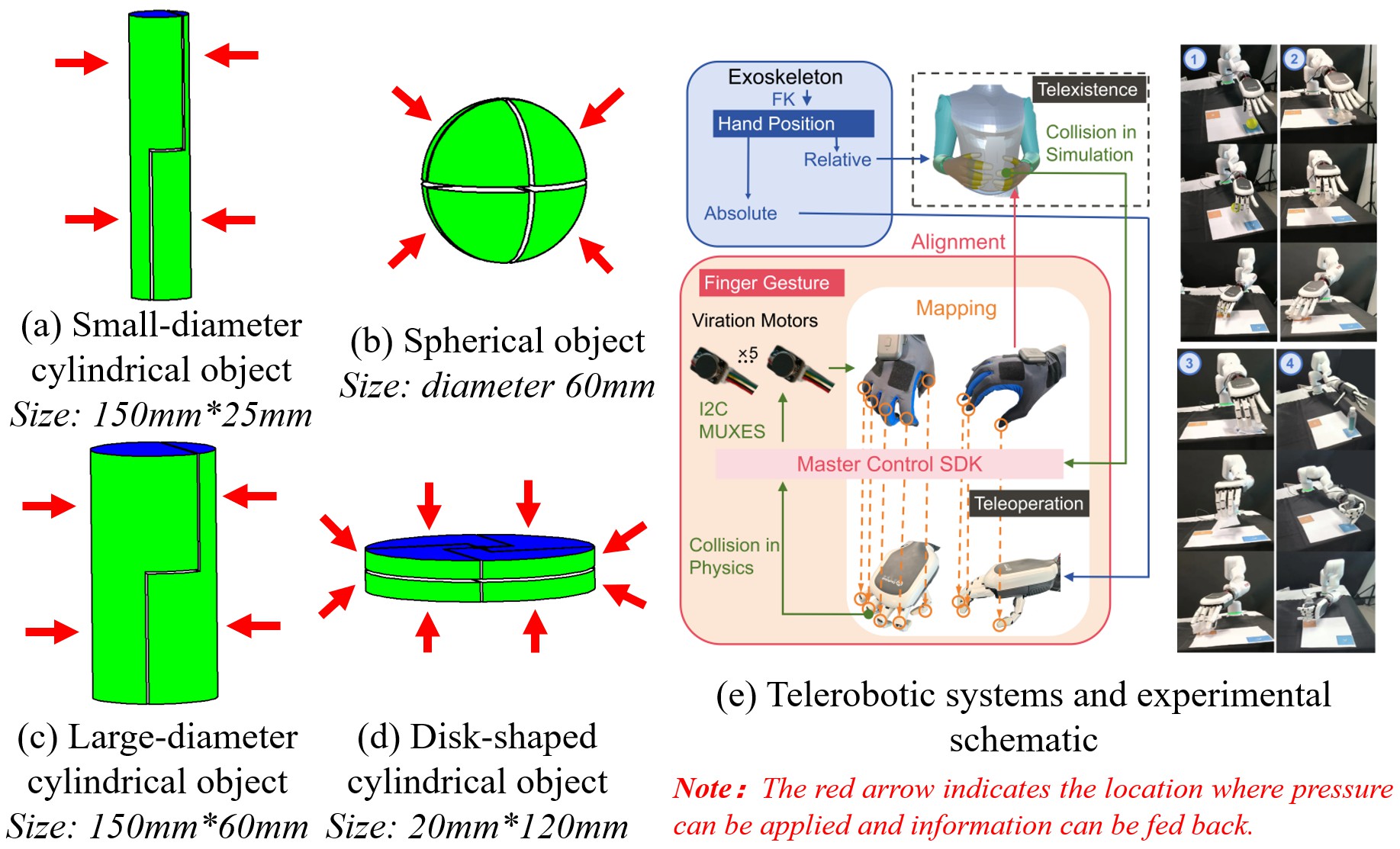}
        \captionsetup{font=footnotesize, labelfont=footnotesize} 
	\caption{Special operated objects equipped with force sensors.}
\end{figure}

\begin{figure*}[t]
        \centering
        \includegraphics[width=0.9\linewidth]{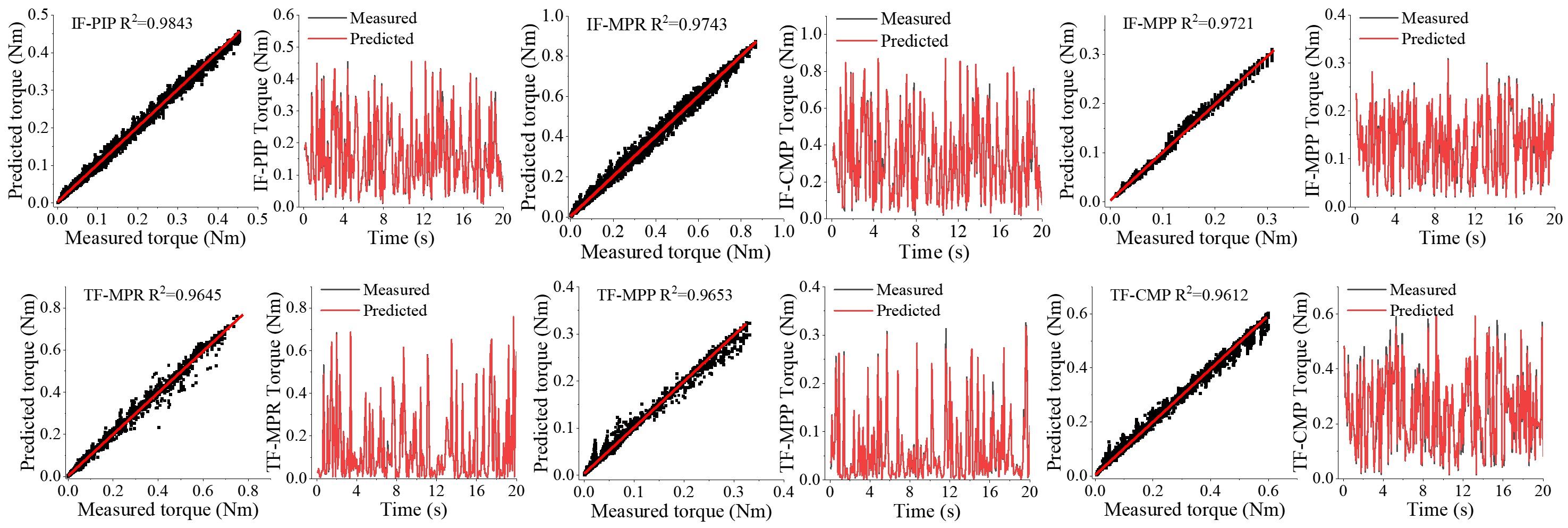}
        \captionsetup{font=footnotesize, labelfont=footnotesize} 
        \caption{Joint torque prediction results.}
\end{figure*}

\begin{figure}[t]
	\centering
	\includegraphics[width=0.9\linewidth]{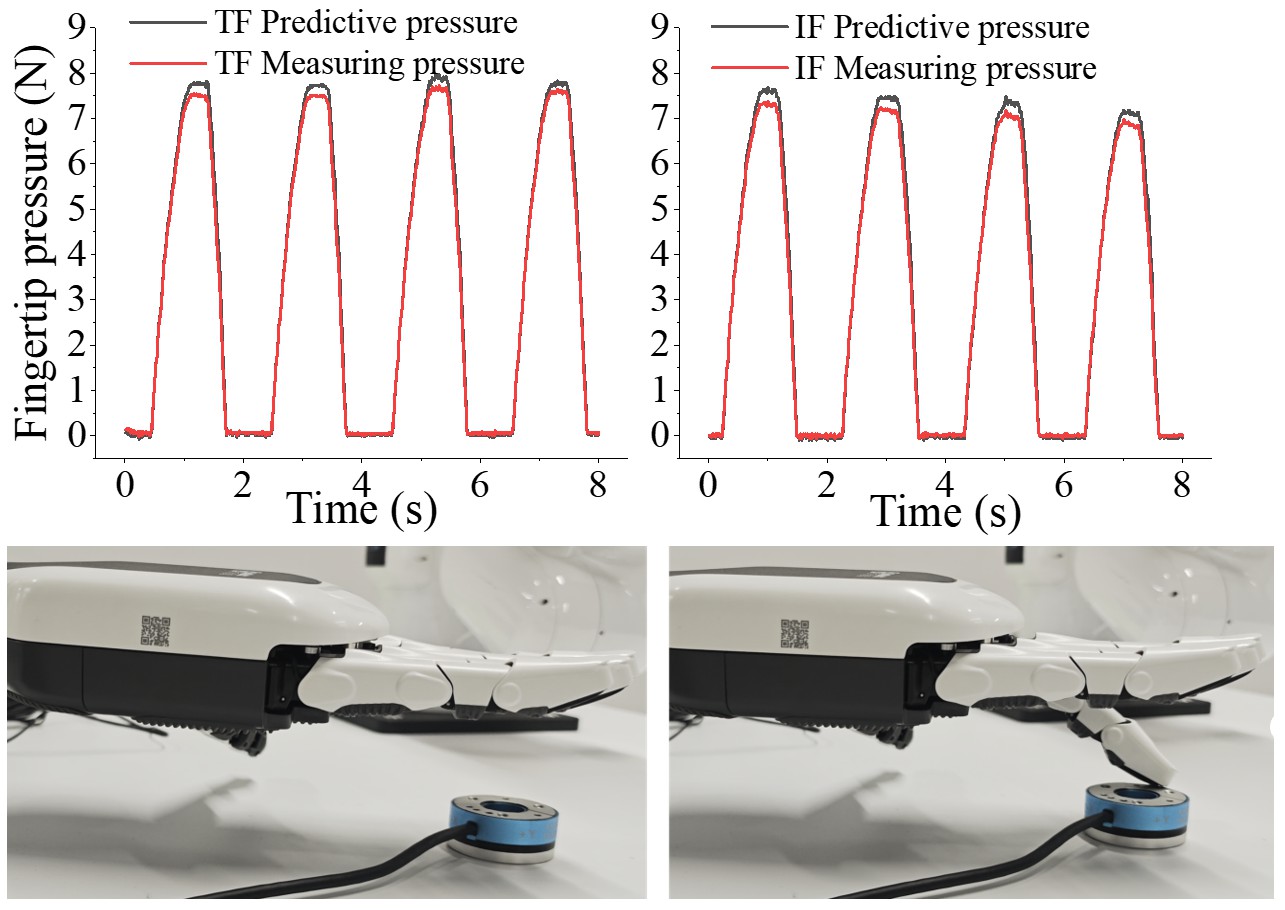}
        \captionsetup{font=footnotesize, labelfont=footnotesize} 
	\caption{Joint torque experimental results.}
\end{figure}

\begin{figure*}[t]
        \centering
        \includegraphics[width=0.9\linewidth]{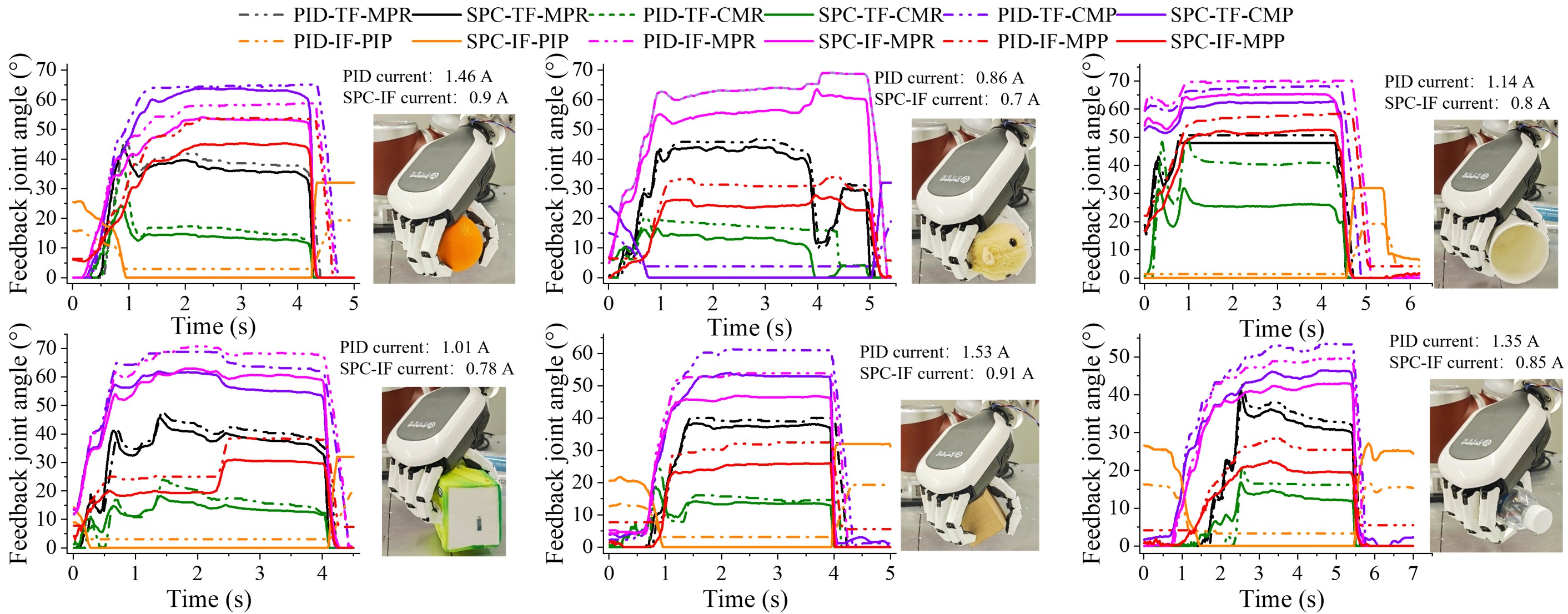}
        \captionsetup{font=footnotesize, labelfont=footnotesize} 
        \caption{The grasping experiments of different hardness objects are based on Self-perceived control and PID.}
\end{figure*}

\begin{figure*}[t]
        \centering
        \includegraphics[width=0.9\linewidth]{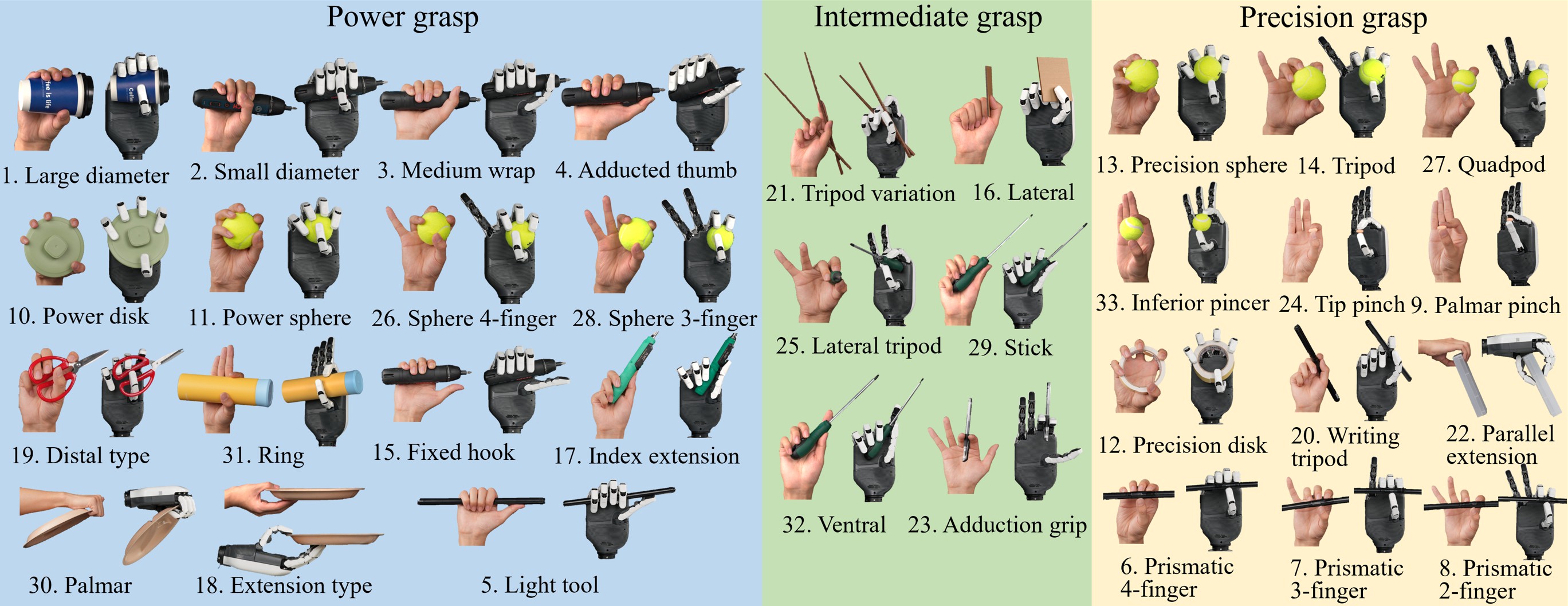}
        \captionsetup{font=footnotesize, labelfont=footnotesize} 
        \caption{The poses and their corresponding indices are in accordance with the GRASP taxonom.}
\end{figure*}

\begin{figure}[t]
	\centering
	\includegraphics[width=0.9\linewidth]{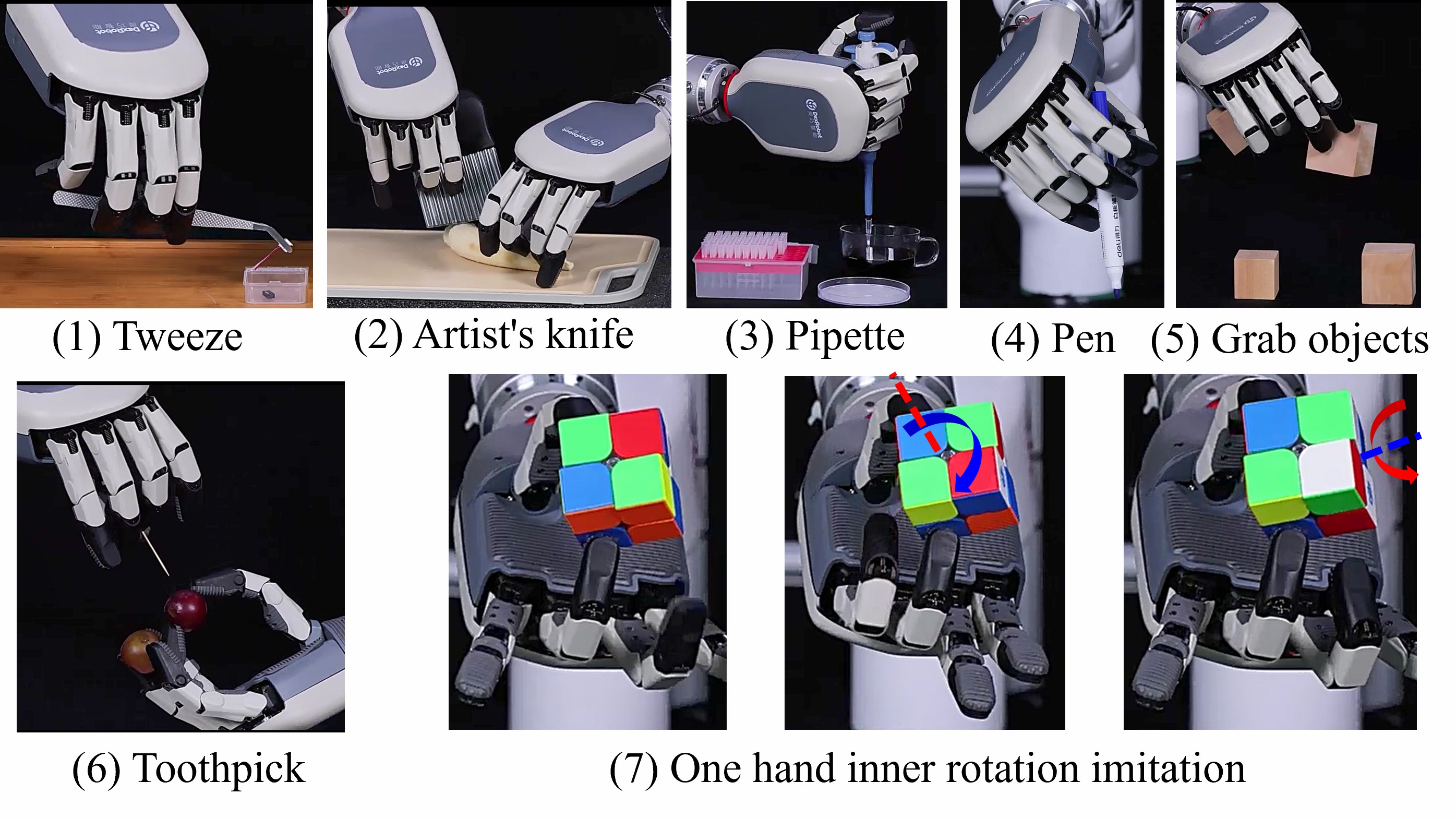}
        \captionsetup{font=footnotesize, labelfont=footnotesize} 
	\caption{Dexterous operation experiment.}
\end{figure}
\begin{table*}[t]
\centering
\caption{Comparison Results of Dexterous Robotic Hand Parameters}
\label{tab:comparison}
\begin{threeparttable}
\resizebox{\linewidth}{!}{%
\begin{tabular}{lccccccccccccc}
\textbf{Object} & \textbf{Human-like} & \textbf{DOF} & \textbf{Active DOF} & \textbf{Load/Weight} & \textbf{Operating speed} & \textbf{Finger load} & \textbf{Force sensing} & \textbf{Position sensing} & \textbf{Running temperature} & \textbf{Communication} & \textbf{Cost} & \textbf{Manufacturability} & \textbf{Durability} \\
\midrule
Human-Hand & 5 & 22 & 22 & 1/10kg & 360°/s & >30 N & 0.1 N & \textbackslash & \(<\)40°C & \textbackslash & \textbackslash & \textbackslash & \textbackslash \\
\textbf{DexHand 021} & 5 & 19 & 12 & 1/5kg & 200°/s & 10 N & 0.1 N & $\pm1$ mm & \(<\)70°C & \makecell{CANFD/EtherCAT/\\ROS/Python/C++} & 9.6 w & 5 & 50W \\
Shadow-Hand & 5 & 24 & 20 & 4.3/4kg & \textbackslash & 10 N & 0.005 N & \textbackslash & \(>\)90°C & \makecell{Ether CAT/CAN} & 100 w & 1 & 1W \\
SVH-Hand & 3 & 20 & 9 & 1/3kg & \textbackslash & 10 N & \textbackslash & \textbackslash & \(>\)90°C & 485 & 65 w & 2 & \textbackslash \\
DLR-Hand & 4 & 15 & 15 & 1.5/5kg & 500°/s & 7 N & 0.01 N & $\pm0.5$ mm & \(>\)90°C & 485 & 70 w & 1 & 10W \\
X-Hand1 & 3 & 12 & 12 & 1/5kg & 360°/s & 15 N & 0.5 N & $\pm0.2$ mm & \(>\)90°C & \makecell{485/\\Ether CAT} & 10 w & 3 & 100W \\
DexH13 & 3 & 16 & 13 & 1.2/5kg & 60°/s & 15 N & 0.1 N & $\pm0.5$ mm & \(>\)90°C & \makecell{Ether CAT/Modbus} & 10 w & 3 & 100W \\
Allegro-Hand & 2 & 12 & 12 & 1/3kg & 500°/s & 6 N & 0.1 N & $\pm0.5$ mm & \textbackslash & Ether CAT & 20 w & 4 & \textbackslash \\
RY-H1 & 3 & 20 & 15 & 1/3kg & 360°/s & 27 N & \textbackslash & \textbackslash & \(>\)70°C & \makecell{485/\\Ether CAT} & 7 w & 4 & 20W \\
ROH-LiteS & 3 & 11 & 6 & 0.7/3kg & 360°/s & 15 N & \textbackslash & \textbackslash & \(<\)70°C & 485 & 5 w & 5 & 20W \\
RHS6DFX & 3 & 11 & 6 & 0.7/3kg & 570°/s & 15 N & 0.5 N & $\pm0.2$ mm & \(<\)70°C & 485 & 5 w & 5 & 20W \\
\bottomrule
\end{tabular}%
}
\begin{tablenotes}
\item[1] ``Human-like'' and ``Manufacturability'' are rated on a scale of 1 to 5, with 5 indicating the optimal parameter.\\
 ``\textbackslash'' indicates the absence of publicly available data.
\end{tablenotes}
\end{threeparttable}
\end{table*}
The mathematical model of admittance control is represented as a mass-damper-spring system, describing the dynamic response of the dexterous hand's joints:
\begin{equation}
    \tau_{\text{ext}} + \tau_d = M_d (\ddot{q}_d - \Delta \ddot{q}) + B_d (\dot{q} - \Delta \dot{q}) + K_d (q - \Delta q) \tag{6}
\end{equation}
where, \(\tau_{\text{ext}}, \tau_d\) are the desired joint torque and load joint torque, respectively;  \(M_d\) is the desired inertia matrix; \(B_d\) is the desired damping matrix; \(K_d\) is the desired stiffness matrix;  \(q_d, \dot{q}_d, \ddot{q}_d\) are the desired joint angle, angular velocity, and angular acceleration, respectively;  \(\Delta q_d, \Delta \dot{q}_d, \Delta \ddot{q}_d\)  are the adjusted joint angle, angular velocity, and angular acceleration under external forces.

In the actual calculation, the discretization control is obtained:

\begin{equation}
    \begin{aligned}
        & \ddot{q} = M_d^{-1} (F_{\text{ext}} - B_d \dot{q} - K_d q) \\
        & \dot{q}(t + \Delta t) = \dot{q}(t) + \ddot{q} \Delta t \\
        & q(t + \Delta t) = q(t) + \dot{q}(t) \Delta t
    \end{aligned}
    \tag{7}
\end{equation}
where, \(q = q_d + \Delta q, \quad \dot{q} = \dot{q}_d + \Delta \dot{q}, \quad \ddot{q} = \ddot{q}_d + \Delta \ddot{q}\) , position control is implemented using PID control, as illustrated in Fig.10.

\section{EXPERIMENTAL VALIDATION}
\subsection{DexHand 021 Performance Experiment}

This study presents a comprehensive performance evaluation of the DexHand 021 robotic hand, focusing on its position control, sensing capabilities, and load capacity, as Fig.11. Experimental results demonstrate the system's robust thermal management, with the hand maintaining a stable internal temperature of 65 \(°C\) during 24-hour continuous measurements using an optical capture system (CHINGMU MC4000W) revealed an absolute positioning error of \(±\)0.002 \(m\) (\(\sigma\)=0.00018 \(m\)) and repeatability error of 0.001 \(m\) (\(\sigma\) =0.0001 \(m\)). The integrated SRI-X-M3703C force sensor achieved \(<\) 0.1 \(N\) measurement error within its 20 \(N\) range, offering advantages in compactness, precision, and integration flexibility compared to alternative solutions ~\cite{10,15}. Load testing confirmed a payload capacity of 1 \(kg\) per finger and 5.5 \(kg\) for the full hand. Benchmarking against state-of-the-art dexterous hands (Shadow-hand, DLR-hand) across 13 performance metrics (Tab.2) highlights the DexHand 021's superior overall capabilities. The DexHand 021 anthropomorphic robotic hand delivers superior manipulation performance with excellent load capacity and thermal management, offering cost-effective solutions while allowing future upgrades in flexibility and sensing.

\subsection{DexHand 021 Performance Experiment}

Limited space precludes torque sensor installation in finger joints. This study employs external calibration by applying known weights to measure single joint bending. motor current, motor position, motor velocity, joint position, joint velocity, temperature, and standard weight data are recorded. The high-precision joint angle sensor provides feedback on finger position and counterweight motion, enabling accurate joint torque measurements. Joint torque is calculated using specified equations (8-14), with data processed offline per Section 3.1 for control validation in Section 3.2. The experimental setup as Fig.12. Control parameters are configured to ensure the weights move at a constant velocity. The external torque \(\tau_{\text{ext}}\)  is calculated using the following formula, where the length \(\overline{BC}\)  in the triangle \(\Delta ABC\)  is determined using the law of cosines:
\begin{equation}
    \angle BAC = \angle_{IF}^{MPF} - \theta_{IF}^{\text{anchor}} + \theta_{IF}^{\text{pulley}} \tag{8}
\end{equation}

\begin{equation}
    \overline{BC} = \sqrt{\left(l_{IF}^{\text{pulley}}\right)^2 + \left(l_{IF}^{\text{anchor}}\right)^2 - 2 l_{IF}^{\text{pulley}} l_{IF}^{\text{anchor}} \cos(\angle BAC)} \tag{9}
\end{equation}

\begin{equation}
    \angle BCA = \arcsin\left(\frac{\overline{BC} \sin(\angle BAC)}{l_{IF}^{\text{pulley}}}\right) \tag{10}
\end{equation}

in the triangle \(\Delta BCE\), the following relationship is obtained based on the law of cosines:
\begin{equation}
    \angle BCE = \arcsin\left( \frac{\overline{BC} \sin\left( \frac{\pi}{2} \right)}{r^c} \right) \tag{11}
\end{equation}
in the triangle  \(\Delta ACD\), the angle between the cable and the finger generatrix is calculated as:
\begin{equation}
    \angle ADC = \angle ACB + \angle BCE - \theta_{IF}^{\text{anchor}} \tag{12}
\end{equation}

the moment arm \(\overline{AD}\) of the cable's pulling force is:
\begin{equation}
\overline{AD} = \sin(\angle ACD) \frac{l_{IF}^{\text{anchor}}}{\sin(\angle ADC)} \tag{13}
\end{equation}

A teleoperation setup was used to perform 31 actions from the GRASP taxonomy, excluding those requiring scissors and chopsticks, which are unfeasible with sensor-equipped props. Sensor-instrumented objects with integrated force sensors were utilized to gather comprehensive and realistic operational data, addressing the limitations of single-joint data acquisition. Grasping action data were collected through teleoperation methods. The objects handled include spheres, large and small diameter cylinders, and disk-shaped items, all equipped with internal force sensors. Fig.13 illustrates the equipment and teleoperation methods used for data collection.

\subsection{Joint Torque Estimation Validation Experiment}

This study investigates torque estimation for the six joints of a modular five-finger system, where the IF configuration is applied to all four fingers. Ground truth data collection experiments were conducted for six different joints. For each joint, loads ranging from 0.05 \(kg\) to 1 \(kg\) were applied, with experiments performed at increments of 0.05 \(kg\). Each load condition was tested for a duration of 30 seconds at a sampling frequency of 100 \(Hz\). In teleoperation grasping experiments, each action was repeated 30 times, with each trial lasting 5 seconds. The experimental temperature range is 20 \(\,^{\circ}\mathrm{C}\) to 50 \(\,^{\circ}\mathrm{C}\), and it is carried out at a gradient of 10 \(\,^{\circ}\mathrm{C}\). This methodology resulted in a total of 1605,000 sample groups. Each sample group contains the information described in Section 3.1 of this paper. The estimation method previously described in Section 4.2 was employed for this analysis. Verification results indicate mean square errors (MSEs) of 0.0006 \(Nm\), 0.0005 \(Nm\), 0.0005 \(Nm\), 0.0006 \(Nm\), 0.0061 \(Nm\), and 0.0061 \(Nm\) for IF-PIP, IF-MPR, TF-MPP, TF-MPR, TF-CMR, and TF-CMP, respectively. The corresponding coefficient of determination \((R²)\) values are 0.9843, 0.9743, 0.9721, 0.9645, 0.9653, and 0.9612, as shown in Fig.14. Thumb shows slightly lower accuracy due to complex cable routing. Further validation was conducted through a fingertip force experiment using a six-axis force sensor (model SRI-X-M3703C). This experiment yielded mean force estimation errors of 0.192 \(N\) for the index finger and 0.15 \(N\) for the thumb finger, with peak errors of 0.899 \(N\) and 0.729 \(N\) at target displacement. The index/thumb finger MSEs are 0.021/0.022 (Fig.15). Incorporating dynamic grasping experiments has enhanced the accuracy of predicting thumb joint movements. This improvement is due to the complex transmission mechanism, where dynamic data helps in better identifying friction parameters. Static estimation errors are larger due to static friction in motor-gearbox/cable transmission, causing significant pulse overshoot. Experiments demonstrate the proposed torque estimation method's importance for improving dexterous hand force sensing. This study evaluates an admittance control method against traditional PID control using DexHand 021. Experiments involved grasping objects of varying hardness, size, and shape (wooden block, simulated fruit, water bottle, paper cup, tissue pack, and plush toy), with 30 trials per object for statistical reliability, Fig.16. The admittance controller, utilizing joint torque estimation, automatically stops motor output upon contact to ensure safe grasping. Results show that while both methods achieve stable grasps, PID control requires significantly larger joint angles. Quantitative analysis revealed that the proposed method reduced joint energy consumption by 40.52 \(\%\), 38.36 \(\%\), 37.04 \(\%\), 29.82 \(\%\), 22.77 \(\%\), and 18.6 \(\%\) respectively for objects ranging from hard to soft, with an average reduction of 31.19 \(\%\). The proposed method provides precise force feedback estimation and dynamic adjustment, enhancing safety, stability, and control accuracy while reducing energy consumption and extending hardware lifespan. These benefits are particularly evident in high-frequency operations and interactions with rigid environments.

\subsection{Dexterous Manipulation Capability Experiment}

In experiments with DexHand 021, a thumb opposition task was compared to human hand movements, demonstrating its ability to replicate human finger flexibility with precision and adaptability. The thumb opposition function, vital for tasks like pinching and grasping, was thoroughly analyzed, highlighting DexHand 021’s potential in complex grasping, robotic manipulation, and human-robot collaboration. Based on the GRASP taxonomy ~\cite{18}, DexHand 021 replicated 33 hand postures, excelling in both "power grasp" (e.g., gripping tools, lifting heavy objects) and "precision manipulation" tasks (e.g., picking pills, holding pens) (Fig.17). Additional tests showcased its capability in tasks like using tweezers, cutting bananas, operating a pipette, writing, and solving a Rubik's Cube, as well as collaborative actions like skewering grapes (Fig.18). These achievements stem from its precise joint control and kinematic design, enabling accurate replication of human hand movements. The results underscore DexHand 021’s potential in robotics, medical rehabilitation, and precision manufacturing, particularly in high-precision, flexible tasks.

\subsection{CONCLUSION}

This paper presents DexHand 021, a highly dexterous robotic hand featuring rope-driven actuation and proprioceptive admittance control. The system showcases impressive capabilities, including a 10 \(N\) single-finger payload, millimeter-level repeat positioning accuracy (0.001 \(m\)), and precise force estimation with an error of less than 0.2 \(N\). These results are based on data from single-finger suspended weight tests and teleoperation grasping experiments. Experimental results show a 31.19 \(\%\) torque reduction compared to conventional PID control during multi-object grasping. The hand successfully executes 33 human-like grasps and complex manipulation tasks ranging from precision tool use to bimanual coordination. Future integration of tactile sensing and embodied intelligence will further enhance its operational capabilities.

\addtolength{\textheight}{-12cm}   



\end{document}